\documentclass[conference]{IEEEtran}
\IEEEoverridecommandlockouts

\usepackage{cite}
\usepackage{url}
\usepackage{amsmath,amssymb,amsfonts}
\usepackage{algorithmic}
\usepackage{graphicx}
\usepackage{textcomp}
\usepackage{xcolor}

\usepackage{multirow}
\usepackage{array}
\usepackage{float}

\def\BibTeX{{\rm B\kern-.05em{\sc i\kern-.025em b}\kern-.08em
    T\kern-.1667em\lower.7ex\hbox{E}\kern-.125emX}}
\begin{document}

\title{AI-Powered Augmented Reality for Satellite Assembly, Integration and Test\\
}

\author{\IEEEauthorblockN{1\textsuperscript{st} Álvaro Patrício}
\IEEEauthorblockA{\textit{Institute For Systems and Robotics} \\
\textit{University of Lisbon}\\
Lisbon, Portugal \\
alvaro.felipe@tecnico.ulisboa.pt}
\and

\IEEEauthorblockN{2\textsuperscript{nd} João Valente}
\IEEEauthorblockA{\textit{Institute For Systems and Robotics} \\
\textit{University of Lisbon}\\
Lisbon, Portugal \\
joao.f.valente@tecnico.ulisboa.pt}
\and

\IEEEauthorblockN{3\textsuperscript{rd} Atabak Dehban}
\IEEEauthorblockA{\textit{Institute For Systems and Robotics} \\
\textit{University of Lisbon}\\
Lisbon, Portugal \\
adehban@isr.tecnico.ulisboa.pt}
\and

\IEEEauthorblockN{5\textsuperscript{th} Inês Cadilha}
\IEEEauthorblockA{\textit{Lusospace} \\
Lisbon, Portugal \\
icadilha@lusospace.com}
\and

\IEEEauthorblockN{5\textsuperscript{th} Daniel Reis}
\IEEEauthorblockA{\textit{Lusospace} \\
Lisbon, Portugal \\
dreis@lusospace.com}
\and

\IEEEauthorblockN{6\textsuperscript{th} Rodrigo Ventura}
\IEEEauthorblockA{\textit{Institute For Systems and Robotics} \\
\textit{University of Lisbon}\\
Lisbon, Portugal \\
rodrigo.ventura@isr.tecnico.ulisboa.pt}
}


\maketitle

%


\begin{abstract}

The integration of Artificial Intelligence (AI) and Augmented Reality (AR) is set to transform satellite Assembly, Integration, and Testing (AIT) processes by enhancing precision, minimizing human error, and improving operational efficiency in cleanroom environments. This paper presents a technical description of the European Space Agency's (ESA) project "AI for AR in Satellite AIT," which combines real-time computer vision and AR systems to assist technicians during satellite assembly. Leveraging Microsoft HoloLens 2 as the AR interface, the system delivers context-aware instructions and real-time feedback, tackling the complexities of object recognition and 6D pose estimation in AIT workflows. All AI models demonstrated over 70\% accuracy, with the detection model exceeding 95\% accuracy, indicating a high level of performance and reliability. A key contribution of this work lies in the effective use of synthetic data for training AI models in AR applications, addressing the significant challenges of obtaining real-world datasets in the highly dynamic satellite environments, as well as the creation of the Segmented Anything Model for Automatic Labelling (SAMAL), which facilitates the automatic annotation of real data, achieving speeds up to 20 times faster than manual human annotation. The findings demonstrate the efficacy of AI-driven AR systems in automating critical satellite assembly tasks, setting a foundation for future innovations in the space industry.

\end{abstract}

\begin{IEEEkeywords}
Artificial Intelligence (AI), Augmented Reality (AR), Space Industry, Computer Vision, Synthetic Data, Microsoft Hololens 2.
\end{IEEEkeywords}

\section{Introduction}

In recent years, the convergence of Artificial Intelligence and Augmented Reality has created new opportunities for innovation across multiple industries, with the space sector being a prime candidate for significant advancements \cite{ai4ar1}. The European Space Agency (ESA) has launched a project titled "AI for AR in Satellite Assembly, Integration, and Testing (AIT)," which aims to enhance the accuracy, efficiency, and automation of satellite production processes. Given the high-stakes nature of satellite manufacturing, minimizing human error during AIT processes within controlled environments, such as clean rooms, is critical to avoiding costly failures that could lead to multi-million-euro losses \cite{error}. 

\begin{figure}[ht]
  \centering
  \includegraphics[width=\linewidth]{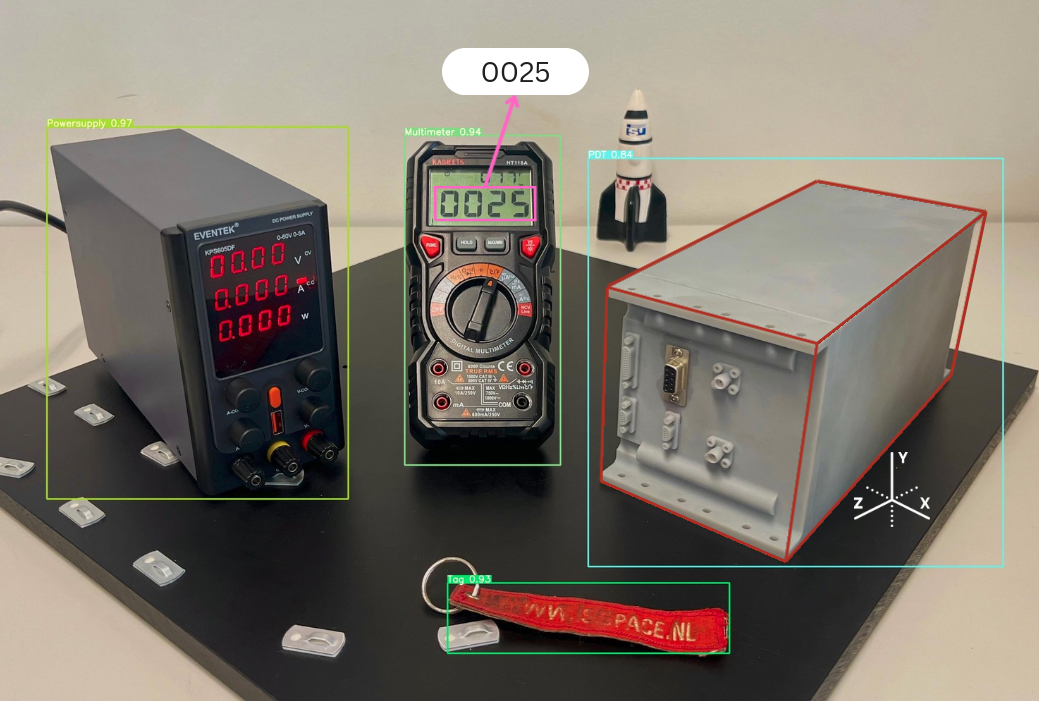} 
  \caption{Demonstrative outputs of the AI-driven augmented reality system for satellite assembly procedures include color-coded bounding boxes: yellow, green, and blue indicate object detection, the red 3D bounding box denotes 6D pose identification, and the pink bounding box represents OCR for measurement instruments.}
  \label{fig:exemplo}
\end{figure}

This project explores the potential of AI-driven real-time object detection systems integrated with AR to assist technicians during satellite assembly. The system is designed to ensure proper component identification and the elimination of unwanted objects within the workspace. Additionally, the project addresses the challenge of generating training data for AI models due to the frequent changes in Satellite components by employing synthetic data. Collaborating organizations, including Luxspace \cite{luxspace}, OHB \cite{ohb}, Lusospace\cite{lusospace}, and Instituto Superior Técnico (IST), contribute their expertise to develop an AI and AR system built around the Microsoft Hololens 2\cite{hololens}, which provides real-time, context-aware instructions to operators while recording relevant data during AIT procedures.

\begin{figure*}[htbp]
  \begin{center}
    \includegraphics[width=0.8\linewidth]{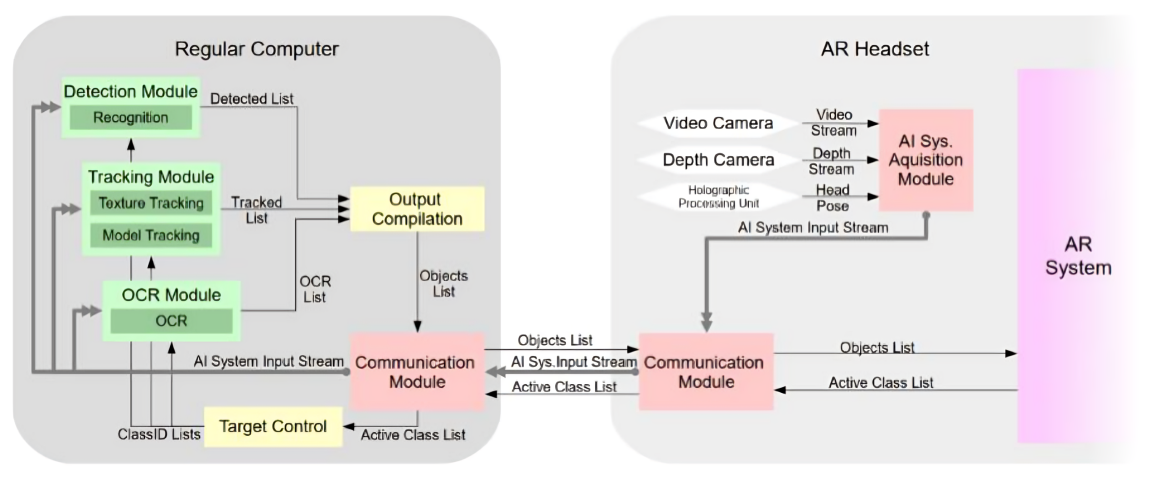}  
  \end{center}
  \caption{The system architecture consists of green modules for processing on the computer, red blocks for communication, and a pink block in the headset for acquiring input data. Gray arrows represent data streams, while black arrows indicate internal data flows.}
  \label{fig:system}
\end{figure*}

The design of the AI system encompasses the entire data processing pipeline, from the capture of RGB, depth, and inertial sensor data through the Hololens 2, to the deployment of advanced computer vision algorithms for object detection, 6D pose, and alphanumeric optical character recognition (OCR). A notable technical challenge is the system’s reliance on synthetic data for training 6D pose models—an approach necessary due to the constant evolution of satellite components and the impracticality of manually labeling real-world datasets.

Several projects across various domains illustrate the potential of AI and AR integration. In collaboration with Microsoft, NASA developed a project called Sidekick that uses Microsoft HoloLens to provide AR-based guidance to astronauts aboard the International Space Station, assisting them in complex procedures by overlaying virtual objects and real-time instructions \cite{nasa}. In medicine, recent advancements in brain surgery techniques have leveraged AI and AR to create 3D holographic models of patient brains, enabling surgeons to visualize pathology with unprecedented clarity and precision during operations, leading to improved outcomes in treating complex neurological conditions \cite{medicine}. These examples demonstrate the versatility of AI-AR systems across industries, from space and medicine to education and cultural heritage.



The key contributions of this paper are threefold. First, we present a novel systems engineering approach specifically developed for the integration of AI and AR in satellite AIT, marking a pioneering effort in the field. This includes the creation of a new, end-to-end system designed to operate offline, addressing the unique technical and operational challenges of this context. Second, we study the effective use of synthetic data for training AI models within AR environments, offering a practical solution to the data scarcity issues faced in highly controlled satellite environments. Finally, the development of the Segmented Anything Model for Automatic Labeling (SAMAL) improves the system's efficacy by enabling the automatic annotation of real data, making the process up to 20 times faster. These contributions lay the groundwork for future advancements in AR applications and satellite assembly processes, with broader implications for the space industry.


\section{Components Overview}

The AI4AR system consists of two key components, as illustrated in Figure \ref{fig:system}: the regular computer and the AR headset. Together, these components work in harmony to enable text and object detection, recognition, real-time feedback, and seamless interaction within a workstation unit environment designed for complex assembly tasks, such as satellite assembly.

\subsection{\textbf{Computer}}

The regular computer handles the bulk of the computational tasks involved in detecting, tracking, and interpreting objects within the environment. Key modules within this component include:

\subsubsection{Detection Module}
 utilizes state-of-the-art object detection algorithms to identify and locate objects or parts within the assembly environment. These detections are used to inform subsequent actions and instructions provided to the user via the AR headset.

\subsubsection{6D Pose Estimation Module}
This module is responsible for determining the 6D pose (position and orientation) of objects and the AR headset in real-time, requiring both rapid and highly precise measurements, often to the millimeter level. This precision and speed are crucial for effective AR headsets, where even minor inaccuracies or delays can significantly impact the user experience by affecting the alignment of virtual elements with the real world.

\subsubsection{OCR  Module}
this component extracts measurements and text from labels, instruments, or other text-based materials present in the assembly scene. It processes text and number data to assist in validation, error detection, or for providing relevant assembly instructions.

\subsubsection{Communication Module}
ensures a smooth data flow between the regular computer and the AR headset. This is accomplished through communication protocols such as ZeroMQ, enabling fast, real-time data transmission necessary for guiding users and processing feedback \cite{zeromq}.


\subsection {\textbf{AR Headset}}
\begin{figure}[]
 \centering
 \includegraphics[width=0.8\linewidth]{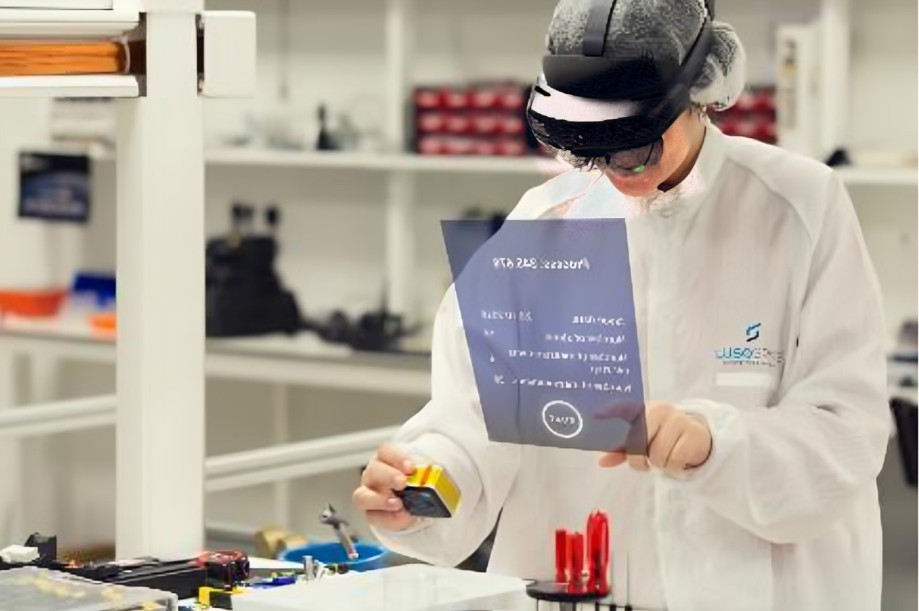} 
 \caption{Illustrative image of a technician utilizing the headset during operational procedures.}
 \label{fig:worker}
\end{figure}

The AR headset serves as the user interface, projecting virtual instructions and augmentations onto the physical assembly environment, Figure \ref{fig:worker} illustrates how a technician views the information on the device. The key components of the AR headset include:

\subsubsection{Video Camera and Depth Camera}
 these cameras capture real-time visual and depth data from the assembly environment, providing essential information for object detection, 6D pose estimation, and scene interpretation.

\subsubsection{Head and Eye Pose Tracking}
 Tracks the position and orientation of the user's head, allowing the virtual overlays to move naturally with the user’s perspective, ensuring proper alignment and consistency as the user navigates the assembly environment.

\subsubsection{Holographic Processing Unit}
Renders 3D holographic overlays into the user’s field of view. This unit provides visual instructions, such as highlighting assembly steps or showing the placement of parts, directly within the user's augmented reality display.

The Communication Module enables efficient real-time data exchange between the computer and AR headset, ensuring synchronized object detection, pose estimation, and virtual instruction projection. Its low-latency communication is critical for the system to respond instantly to user movements during assembly.

\section{Literature Review}

\subsection{Object Detection}

Over the past decade, object detection has evolved from traditional techniques reliant on handcrafted features, such as Haar cascades and Histograms of Oriented Gradients (HOG)\cite{HOG}, to advanced deep learning models \cite{odevolution}. The breakthrough came with Convolutional Neural Networks (CNNs), highlighted by AlexNet’s success in 2012 \cite{alexnet}, which demonstrated the effectiveness of deep learning in image classification. Subsequent models, such as Region-based CNN (R-CNN)\cite{rcnn} and YOLO (You Only Look Once) \cite{yolo}, revolutionized object detection by enabling automatic feature learning and real-time, accurate detection. This transition has greatly enhanced the robustness and scalability of object detection solutions across various applications.

YOLOv7 was chosen for this project due to its exceptional balance of speed and accuracy in object detection, making it ideal for real-time applications such as augmented reality. Its advanced architecture, which builds on the strengths of previous YOLO versions, includes improvements like "Bag of Freebies" for data augmentation and multi-head detection, which enhance precision without sacrificing efficiency. This makes YOLOv7 highly effective for detecting and localizing objects in complex environments \cite{yolov7}.  

Additionally, YOLOv7's support for transfer learning and compatibility with large datasets like COCO allows for faster training and better generalization across diverse scenarios \cite{COCO}. The model's ability to leverage pre-trained weights and adapt to specific tasks ensures robust performance, even with smaller datasets. Its proven success in industry and research, coupled with its flexibility, made YOLOv7 the optimal choice for achieving high-performance object detection in this project.

\subsection{Accelerating Data Annotation with SAM2}

In the context of object recognition, we employed a novel technique to annotate real-world data with significantly increased efficiency. This approach utilizes SAM2 (Segment Anything Model 2), an advanced segmentation model specifically designed to automatically identify and segment objects across a diverse array of images and videos \cite{sam2_paper}. 

SAM2 operates on the principle of utilizing large-scale training data and sophisticated machine learning algorithms to achieve state-of-the-art performance in object segmentation tasks. It leverages a combination of deep learning techniques and a transformer-based architecture to facilitate the segmentation of objects in varying contexts, lighting conditions, and backgrounds. This versatility allows SAM2 to generalize effectively across different domains, making it particularly valuable in applications requiring rapid and accurate image annotation.

SAM2 automates the labeling process, significantly decreasing the time and effort needed for manual annotation. This automation enhances the quality and consistency of the annotations, which is vital for training effective object detection models and improving performance in real-world recognition tasks.

\subsection{6D pose}

Over the past decade, 6D pose estimation has advanced significantly, transitioning from traditional geometry-based methods, such as template matching and point-pair features (PPF), which relied heavily on manual feature engineering, to sophisticated deep learning models\cite{ppf}. Early techniques struggled to handle complex real-world scenarios with occlusions and lighting variations. The advent of deep learning, particularly with Convolutional Neural Networks (CNNs), transformed the field by enabling models like PoseCNN \cite{posecnn} and DenseFusion \cite{densefusion} to learn robust, data-driven representations that could accurately predict 6D object poses under challenging conditions, offering greater flexibility and precision across diverse environments.   

We selected GDRNPP (Geometrically Disentangled Representation Network with Probabilistic Pose) for our 6D pose estimation task due to its state-of-the-art architecture, which optimizes both precision and computational efficiency. Moreover, GDRNPP supports transfer learning, enabling the model to benefit from pre-trained networks and adapt quickly to new datasets with minimal retraining. This efficiency in leveraging both geometric information and deep-learned features ensures robust performance even with smaller or more diverse datasets \cite{GDRNPP1} \cite{GDRNPP2}. 

Our choice of GDRNPP is particularly suited for augmented reality (AR) applications, where both speed and precision are crucial. In AR, real-time processing is essential for ensuring a seamless user experience, as any delay or inaccuracy in pose estimation can result in poor alignment of virtual objects with the real world, breaking immersion. Additionally, GDRNPP incorporates geometric constraints, which contribute to more accurate pose predictions in scenarios involving partial visibility or overlapping objects, making it a suitable choice for complex, real-world applications.

\subsection{OCR}

For the task of Optical OCR, two open-source solutions were considered: EasyOCR \cite{EasyOCR} and MMOCR \cite{MMOCR}. Tesseract, a widely known OCR engine, was excluded from the benchmark due to its reliance on pre-processed, clean, text-only images and its limitation to horizontal rectangular boxes, which renders it unsuitable for OCR in unconstrained environments as encountered in this study.

The benchmarking of text detection models revealed that EasyOCR demonstrates the highest inference speed (6.47 FPS) and the best performance at the 0.5 F1-score threshold, making it the most suitable choice for real-time applications despite lower accuracy at more stringent thresholds compared to other models. For text recognition, the speed of inference was significantly higher (130.61 FPS), which is crucial given the real-time constraints. This trade-off in accuracy is acceptable considering the need for rapid processing.

Considering all reasons stated above, EasyOCR was identified as the most appropriate model for both text detection and recognition tasks, offering a favorable balance between speed and baseline performance, thus serving as an optimal candidate for further enhancements.


\section{Implementation pipeline components}

This section provides a detailed overview of the individual components used in the project "AI for AR in satellite AIT." Each component contributes to the system’s capability of integrating real-time object detection, computer vision, and AR to assist technicians in satellite assembly tasks.

\subsection{Unity}

Unity serves as the core 3D engine in the AI4AR system, chosen for its powerful and accessible platform for developing augmented reality interfaces on Microsoft HoloLens 2. The system relies heavily on Unity’s capabilities to create high-fidelity 3D visualizations and overlay real-time information, assisting operators in satellite Assembly, Integration, and Testing. \cite{unity}. 

The project utilizes the Mixed Reality Toolkit (MRTK) 3, an open-source toolkit provided by Microsoft, to streamline the development of mixed reality applications. MRTK enables efficient handling of user interactions, hand tracking, and gaze input, allowing for intuitive and responsive user experiences on the HoloLens 2 \cite{
MRTK}. In conjunction with Unity, MRTK provides pre-built components and controls optimized for mixed reality, simplifying the implementation of interactive and immersive interfaces.

Input data from the HoloLens is processed using OpenXR, which integrates seamlessly with MRTK to manage device input such as hand gestures and spatial tracking \cite{openxr}. These inputs enable precise interaction with the virtual elements overlaid in the operator’s field of view.

To handle data communication between Unity and the regular computer, a custom C\# DLL was developed. This DLL enables asynchronous communication, using ZeroMQ to efficiently send requests and receive data such as object detection, OCR, 6D pose estimation, and other AI outputs. The data is then processed by Unity to dynamically update the holographic overlays and provide real-time feedback to the operator, as shown in Figure \ref{fig:system}.

\subsection{Yolov7}

YOLOv7 (You Only Look Once, version 7) is a cutting-edge real-time object detection model, recognized for its high performance across a range of object detection benchmarks \cite{yolov7}. To optimize performance, the project experimented with different pre-trained models in YOLOv7. While some of these models achieved superior precision and recall, they often came at the cost of reduced detection speed. Given the importance of real-time detection for effective AR implementation, this trade-off between speed and accuracy was carefully considered \cite{ai4ar1}. To test your object detection and ensure the reliability of the model in practical applications, we established a dedicated test set captured using the actual AR equipment employed in the assembly process. The test images were manually annotated using Roboflow's labeling tools \cite{roboflow}, providing high-quality ground truth labels for performance evaluation. Metrics such as accuracy, recall, mean Average Precision (mAP) (specifically  mAP@0.5-0.95, and mAP@0.95) were employed to assess and compare the performance of different YOLOv7 models. This rigorous evaluation process ensures that the model selected for deployment offers the best possible balance of speed, accuracy, and robustness in a real-world AR-enhanced Satellite assembly scenario.

\subsection{SAM2}

\begin{figure}[]
  \centering
  \includegraphics[width=\linewidth]{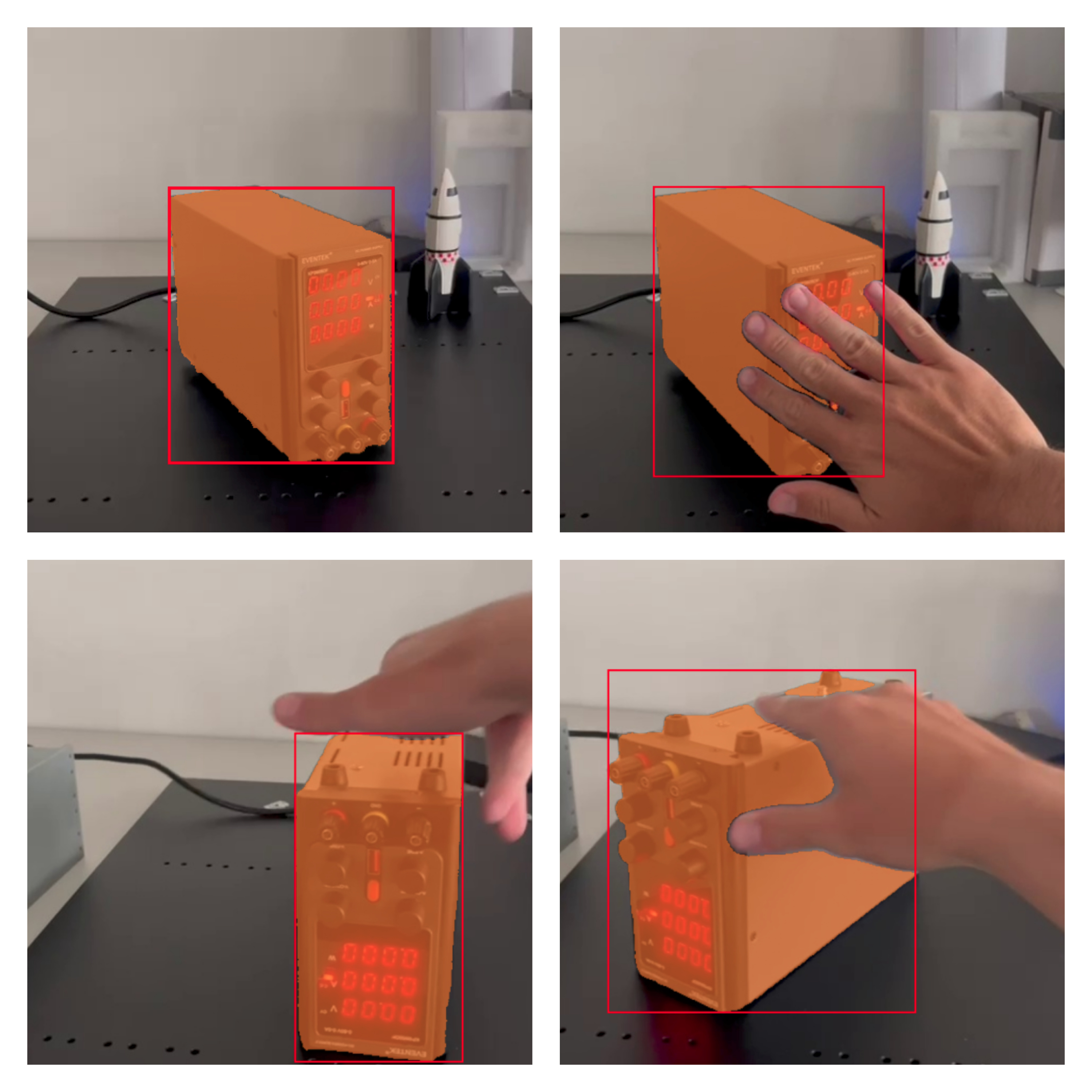} 
  \caption{SAMAL is an advanced annotation tool that accurately generates bounding boxes around objects. It effectively handles occlusions, such as user hands, and adapts to changes in object positions, making it especially valuable for augmented reality applications where hands frequently interact with objects.}
  \label{fig:sam2}
\end{figure}

The integration of SAM2 for automatic annotation enabled a more efficient data preparation pipeline, allowing for the generation of large volumes of annotated images within a matter of minutes. These annotated datasets will subsequently be used to train object detection models, enhancing their performance in real-world satellite assembly applications. According to experimental results, the use of SAM2 for automating the annotation process yielded highly positive outcomes, as you can see in some of the examples in Figure \ref{fig:sam2}, reducing annotation time by several orders of magnitude compared to traditional manual methods, thereby streamlining the dataset preparation phase for model training.

To facilitate automatic labeling, an adaptation of the original SAM2 code was developed to create SAMAL. Initially, a series of recordings were captured from various angles, simulating the diverse perspectives a worker might have of the object in question. This process included variations in distance, lighting conditions and others. Subsequently, the object of interest is selected in the first frame of the video for tracking throughout the entire sequence of frames. SAM2 then generates a mask that delineates the location of the targeted object in all frames. Using these masks, we identify the extreme points in both the X and Y coordinates, from which we derive the bounding box coordinates in the format required to train YOLOv7. 

\subsection{GDRNPP}

GDRNPP (Generalized Differentiable Rendering Network for Pose Prediction) is a state-of-the-art 6D pose estimation method deployed in this project to accurately localize and provide 6D position of Satellite components during the assembly process \cite{GDRNPP1}. GDRNPP was chosen for its ability to provide highly precise pose estimations, which are essential in augmented reality (AR) environments where even a slight deviation can lead to critical errors. For instance, in our specific case, if the model's estimated pose is off by just a few millimeters from the real component, a hole intended for a screw could be misaligned, causing issues during assembly.

The model's ability to operate exclusively with RGB data optimizes the utilization of the HoloLens camera, bypassing the limitations associated with the lower resolution and frame rate of depth sensors while ensuring smooth real-time performance. The performance of GDRNPP in the BOP Challenge 2022 further validated its reliability, establishing it as the most appropriate open-source solution for this project \cite{bop2022}.

To evaluate the system's performance, we use the most widely accepted metrics in 6D pose estimation and tracking: ADD and ADD(-S) for symmetric objects. These metrics rely on the average distance between corresponding points in the 3D model, comparing the estimated and ground-truth poses. An estimated pose is considered correct when the average distance is below a specific threshold, typically set at 10\% of the object's diameter. Accuracy is then calculated as the ratio of correct estimations to the total number of images in the dataset, ensuring that GDRNPP delivers both the precision and reliability required for AR-enhanced Satellite assembly.

To evaluate the GDRNPP framework, we created a custom test dataset with 188 images (640 × 360 resolution) captured using the HoloLens. In one of the scenes, 73 frames were taken while the PDT was rotated in the user’s hands, simulating real-world manipulation and introducing challenges like hand occlusions. The 6D poses were annotated using 6D-PAT \cite{6DPAT} by manually aligning the 3D model to the object in the image and establishing at least six 2D-3D correspondences, which were then used to compute the pose with a PnP solver. This provided accurate ground truth data for performance evaluation.

\subsection{EasyOCR}

EasyOCR is a Python module for extracting text from images, developed by Jaided AI in 2020. It is a general OCR package that can read both natural scene text and dense text in a document, with support for multiple languages. It leverages the CRAFT \cite{CRAFT} algorithm for text detection and a CRNN \cite{CRNN} for text recognition.

In this project, we recognize the critical importance of accurate readings from the display of devices, which influenced our decision to train OCR models. Two OCR models are being developed: one dedicated to recognizing numbers, which has been completed and is currently utilized to detect multimeter and power supply measurements, as shown in Figure \ref{fig:exemplo}, and another designed for recognizing both letters and numbers, specifically for processing tags. This decision of training two different OCR models was made due to the importance of readings from the display of the devices.  It's also important to note that both the text detection and text recognition tasks for the alphanumeric OCR model are actively undergoing modifications.

\begin{table*}[ht]
\centering
\caption{Performance metrics comparison across different classes for the detection module}
\normalsize 
\setlength{\tabcolsep}{10pt} 
\begin{tabular}{|l|l|c|c|c|c|}
\hline
\textbf{Dataset}     & \textbf{Class} & \textbf{Precision} & \textbf{Recall} & \textbf{mAP@0.5} & \textbf{mAP@0.5:0.95} \\ \hline
\multirow{1}{*}{\textbf{Synthetic Data}}   & All        & 0.99  & 0.60  & 0.604 & 0.467 \\ \hline
\multirow{4}{*}{\textbf{Real Data}}        & All        & 0.991 & 0.978 & 0.976 & 0.753 \\ \cline{2-6} 
                                      & Powersupply & 1.0   & 0.960 & 0.965 & 0.828 \\ \cline{2-6} 
                                      & Multimeter  & 1.0   & 0.992 & 0.998 & 0.826 \\ \cline{2-6} 
                                      & Tag         & 0.980 & 0.975 & 0.966 & 0.604 \\ \hline
\end{tabular}
\label{tab:results_comparison}
\end{table*}

\subsubsection{\textbf{Numerical OCR Model}}

With regard to the numerical ocr model, the training datasets for text detection and recognition tasks were synthetically generated to enhance the performance of the EasyOCR models. For the text detection model focused on numbers, 100,000 synthetic samples were created using SynthText \cite{SynthText}. This dataset generation process involved several key steps to create realistic synthetic images for training the OCR system. First, video footage from the test dataset was captured to provide realistic backgrounds that closely mirrored real-world environments. Text was then sampled from a dictionary of four-digit positive and negative numbers and placed on suitable regions of the video frames by analyzing predicted depth and segmentation maps, ensuring the text fit semantically with the scene. The text was rendered in the seven-segment DSEG font to replicate the appearance of LCD displays and blended with the background images, maintaining natural illumination gradients for enhanced realism.

For text recognition, a synthetic dataset of 100,000 samples was generated, covering 12 character classes (“ .-1234567890”). Text crops were created using the Text Recognition Data Generator (TRDG \cite{TRDG}), introducing variations in size, color, noise, and other factors to improve robustness in diverse real-world conditions. Techniques inspired by the MJSynth dataset generation process \cite{MJSynth}, such as random perspective transformations, simulated real-world variations in text orientation. Finally, synthetic images were further enhanced by blending them with patches from the IC15 Scene Text Detection dataset, adding diverse textures and lighting conditions to better simulate natural environments.

\subsubsection{\textbf{Alphanumeric OCR Model}}

Both the text detection and recognition tasks for this model follow a similar approach as the numerical OCR, with the main difference being the classes that represent the characters identified in the test set, which include a broader range of alphanumeric characters beyond numbers.

\section{Experiments and Results}

All benchmarks were conducted on a standard desktop equipped with an AMD Ryzen 5 2600 @ 3.4GHz CPU and an NVIDIA GeForce GTX 1070 GPU. The evaluation metrics included average Frames Per Second (FPS) for inference speed, alongside accuracy measures specific to each task.

\subsection{Object Detection Task:}
Although synthetic data played a crucial role in addressing other AI tasks, it became evident that real data produced better results when compared to synthetic data in object detection. With the development of a tool for automating the annotation of these datasets for training, the use of synthetic data became less favorable. This is because it was possible to generate complete training datasets for various objects in just a few minutes. However, due to the model's low recall (high rate of false negatives), we developed SAMAL as an alternative, efficient method for collecting training data. This allowed us to conduct a comprehensive comparison between SAMAL and the previously generated synthetic data.

To address these challenges and enable more robust model comparisons, we utilized two datasets: one with real data and another with synthetic data. The results of these experiments will be presented, comparing two distinct data generation strategies for training object detection models:

\begin{itemize} \item \textbf{Real Data Only}: This experiment utilized 4,443 images for training and 1,111 images for validation. \item \textbf{Synthetic Data Only}: For this setup, 8,210 images were used for training, and 1,760 images were used for validation. \end{itemize}

For consistency and comparability, all models were trained for 10 epochs using an NVIDIA A100 GPU with 80GB of memory. Additionally, all models were trained using the same pre-trained weights, specifically \texttt{yolov7.pt}. The results for all models were reported using a confidence threshold of 0.5.

For the generation of synthetic data, various approaches were employed, including object size variation, changes in lighting conditions, the inclusion of distractors, and the application of motion blur to simulate rapid camera movements. The model trained exclusively on synthetic data demonstrated good overall precision of 0.99, but with a recall of 0.60 and a mAP@0.5:0.95 score of 0.467, it fell short compared to the real data model, as illustrated in Table \ref{tab:results_comparison}. Conversely, the model trained on real data achieved near-perfect precision of 0.991 and a recall of 0.978, with a significantly higher mAP@0.5:0.95 score of 0.753, as demonstrated in Table \ref{tab:results_comparison}. This highlights the superior performance of real data in capturing the variability and complexity of real-world object detection scenarios.

\begin{figure}[]
  \centering
  \includegraphics[width=\linewidth]{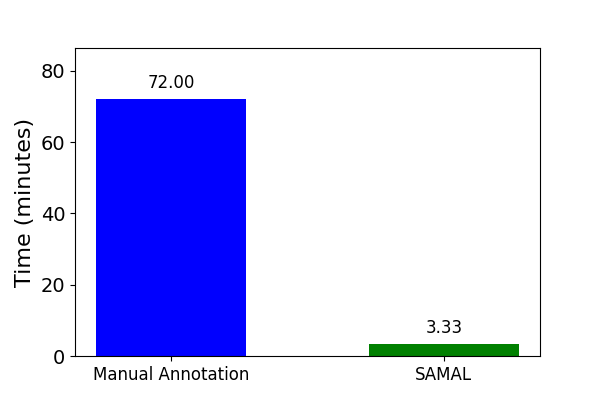} 
  \caption{The chart compares the time required for manual annotation of 459 images versus SAMAL. While manual annotation took 72 minutes, SAMAL completed the same task in just 3 minutes and 20 seconds, demonstrating a substantial increase in efficiency.}
  \label{fig:sam_better}
\end{figure}

Historically, the manual annotation of real data has been a major bottleneck, often consuming hours of labor. For instance, labeling a set of 459 images required a human operator 72 minutes to complete, and even then, the resulting dataset was not perfectly annotated, with many bounding boxes of different sizes than those desired. In our experiments, SAMAL annotated 459 images in just 3 minutes and 20 seconds—more than 20 times faster than manual annotation, as illustrated in Figure \ref{fig:sam_better}. This remarkable speed improvement is accompanied by increased accuracy and consistency, as it minimizes the potential for human error.

SAMAL’s effectiveness in handling occlusions, such as those caused by a user’s hand in AR environments. Since users' hands frequently interact with objects of interest, SAMAL’s ability to accurately generate bounding boxes around objects, even in complex scenarios with occlusions and positional changes, is essential. While synthetic data provides a useful alternative for some object detection tasks, the superior performance of real data remains evident. The primary limitation of real data, which is its time-consuming annotation process, has been largely addressed by the development of SAMAL. This tool transforms the labeling process into an efficient, automated workflow, enabling faster and more accurate dataset preparation. The innovation not only enhances the scalability of real data in object detection but also improves the accuracy and consistency of models, particularly in real-world and AR-enhanced applications.

\subsection{6d Pose Task}

The 6D Pose Tracking model was trained using a modified version of the GDRNPP training script, originally designed for public BOP Challenge datasets \cite{BOP-Datasets}, to accommodate custom data. Key parameter adjustments included setting the "DZI PAD SCALE" to 1 and "COLOR AUG PROB" to 0, preventing background replacement in training images. The training dataset, comprising 50,000 samples, was generated using BlenderProc \cite{BlenderProc}, adhering to the synthetic data creation steps used in the BOP Challenge 2022 \cite{BOP-Datasets}.

With regards to the data generation process, several steps were taken:

\begin{itemize}
    \item Synthetic scenes were created by arranging planes as walls with realistic PBR textures;
    \item Lighting conditions were simulated by randomizing the number of lights in the scene,their respective strength, color and position;
    \item The 3D model of PDT was randomly placed alongside distractor objects with varied materials to create more complex and realistic interactions;
    \item Physics simulations were also carried out to ensured plausible object poses and consistency with reality;
    \item Multiple cameras were used to render RGB images, object masks, and visibility masks with annotated 6D poses.     
\end{itemize}

Furthermore, because the materials used as backgrounds of the scenes rooms were very different from the background present in the test set, rendered PDTs were blended into images of the test environment using visibility masks, preserving occlusion effects and thus minimizing the domain gap between synthetic and real test images.

\begin{figure}[H]
\begin{minipage}[c]{0.25\textwidth}
    \centering
    \includegraphics[width=4.5cm]{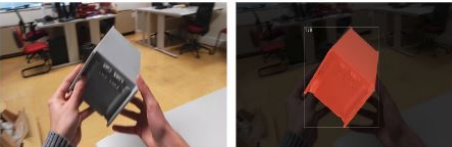}
    \caption{6D Pose Test Scene A}
    \label{6d_sceneA}
\end{minipage}
\begin{minipage}[c]{0.2\textwidth}
    \centering
    \includegraphics[width=4.5cm]{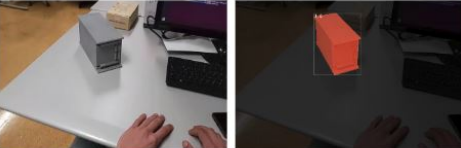}
    \caption{6D Pose Test Scene B}
    \label{6d_sceneB}
\end{minipage}
\end{figure}

The GDRNPP requires an initial bounding box around the object, which was derived from object masks rendered in BlenderProc for training, and from an object detector during test-time. Considering two scenes: scene A (figure \ref{6d_sceneA}), where the PDT is being rotated in the hands of the user, thus inducing strong occlusions with the hands, and scene B (figure \ref{6d_sceneB}), in which the PDT is static on the table and the camera is moving around while looking at it, one can visualize the results after training in Table \ref{table: 6dpose_tracking_test}.

\begin{table}[ht]
\centering
\caption{6D Pose Tracking results on test dataset.}
\small 
\begin{tabular}{|c|c|c|}
\hline
\textbf{Training Dataset}       & \textbf{BlenderProc} & \begin{tabular}[c]{@{}c@{}}\textbf{BlenderProc +}\\ \textbf{Room backgrounds}\end{tabular} \\ \hline
\textbf{Scene A (\%)}           & 66.00       & 73.97                                                                    \\ \hline
\textbf{Scene B (\%)}           & 76.19       & 88.70                                                                    \\ \hline
\textbf{Full test dataset (\%)} & 69.33       & 82.98                                                                    \\ \hline
\textbf{Speed (FPS)}            & 12.92       & 12.92                                                                    \\ \hline
\end{tabular}
\label{table: 6dpose_tracking_test}
\end{table}

One can see that training with blended images showed significant improvements particularly in scene B, which presents fewer occlusions. The model operated at 13 FPS, suitable for the intended application. Accuracy, though lower than OCR models, established a satisfactory baseline for system testing.

To evaluate the impact of using less reliable bounding boxes, the performance of the model was tested using altered bounding boxes to simulate the output of an object detector. For each test sample, the ground-truth bounding box was randomly shifted up to 25\% of its width and height and scaled by a factor between 0.75 and 1.25 of its original size. The average accuracy over five tests on the full dataset is presented in Table \ref{table: altering_bboxes_6dpose}.

\begin{table}[ht]
\centering
\caption{Effect of altering the initial bounding boxes in 6D Pose Tracking results.}
\small 
\begin{tabular}{|c|c|}
\hline
\textbf{Bounding box} & \textbf{Full test dataset (\%)} \\ \hline
\textbf{Ground-truth} & 82.98                  \\ \hline
\textbf{Altered}      & 78.51                  \\ \hline
\end{tabular}
\label{table: altering_bboxes_6dpose}
\end{table}

Despite significant distortions to the bounding boxes, model performance decreased by only about 5\%, which suggests the model's robustness to bounding box inaccuracies.

\subsection{OCR Task}

\subsubsection{\textbf{Numerical OCR}}

The text detection model was trained using EasyOCR’s \cite{EasyOCR} synthetic data training script, which requires character-level annotations. Key configuration parameters were adjusted to optimize performance, including a text threshold of 0.5, a low text value of 0.3, a link threshold of 0.1, a canvas size of 2240, a magnification ratio of 1.5, no additional margin, and a slope threshold of 0.

As mentioned in a previous section, the text detection model was trained using the CRAFT \cite{CRAFT} algorithm on a synthetic dataset. The comparison of the trained model's performance with the pre-trained EasyOCR model can be viewed in Table \ref{table: text_detection_results}.

\begin{table}[ht]
\centering
\caption{Text detection results after training.}
\small 
\begin{tabular}{|c|c|c|c|}
\hline
\textbf{Model}         & \textbf{F1 @ 0.5} & \textbf{F1 @ 0.7} & \textbf{Speed (FPS)} \\ \hline
\textbf{EasyOCR (pretrained)}  & 0.8858            & 0.4890            & 6.47                 \\ \hline
\textbf{Trained model} & 0.9933            & 0.9797            & 2.91                 \\ \hline
\end{tabular}
\label{table: text_detection_results}
\end{table}

The trained model achieved near-perfect F1 scores at both IoU thresholds (0.5 and 0.7), demonstrating substantial improvement over the pre-trained EasyOCR model. This performance was attained by setting the magnification ratio to 1.5, which increases input image size, thereby enhancing detection accuracy but reducing inference speed to 2.91 FPS. This speed, while not real-time, is adequate for applications involving measuring instruments with static or slow-changing displays.

For applications requiring higher frame rates, adjustments to the magnification ratio can be made, providing a balance between accuracy and speed. Lowering the ratio reduces image resolution, impacting detection performance and allowing for flexible optimization based on specific use-case requirements.

With regards to the text recognition model, it was trained using the CRNN algorithm on the generated synthetic dataset and it demonstrated significant improvements compared to the EasyOCR pre-trained model, as one can see in Table \ref{table: text_recog_results}.

The trained model achieved an accuracy of 98.36\%, far surpassing the performance of the pre-trained model. It also exhibited high inference speeds (188.82 FPS), meeting real-time requirements. This increase in speed is partly due to the reduced number of classes (digits, decimal, and negative symbols), which simplified the character classification task. The speed gains were further enhanced by setting a batch size greater than one during inference, which was done because multiple bounding boxes are detected per frame and then sent in order to be recognized.

The combined OCR pipeline, integrating the text detection and recognition models, was evaluated to assess its overall performance in reading numbers. The text detection model identifies bounding boxes in input images, which are then cropped and processed by the recognition model to extract the detected numbers.

The text recognition test dataset was created using ground-truth bounding boxes to compare performance against detected bounding boxes, highlighting the impact of using real versus ideal conditions. The quantitative results are shown in Table \ref{table: text_recog_results}.

\begin{table}[ht]
\centering
\caption{Text recognition Results Comparison}
\small 
\begin{tabular}{|c|c|c|}
\hline
\textbf{Model}                & \textbf{Accuracy (\%)} & \textbf{Speed (FPS)} \\ \hline
\textbf{EasyOCR (pre-trained)} & 2.35                   & 130.61               \\ \hline
\textbf{Oracle OCR}           & 98.36                  & 188.82               \\ \hline
\textbf{OCR Pipeline}         & 96.03                  & -                    \\ \hline
\end{tabular}
\label{table: text_recog_results}
\end{table}

The OCR pipeline achieved an overall accuracy of 96.03\%, reflecting a slight decrease of 2.3\% compared to using ground-truth bounding boxes. Despite this, the combined pipeline remains highly effective, correctly reading numbers in 96\% of cases. These results confirm the robustness of the developed OCR module, demonstrating its potential for accurate numerical reading in practical scenarios.

\subsubsection{\textbf{Alphanumeric OCR}}

As mentioned above, the text detection and recognition tasks for this OCR model are currently undergoing modifications, however a comparison between this model and its corresponding Oracle version (obtained by performing only the text recognition task on top of the ground truth bounding boxes) was made and can be seen in Table \ref{table: oracle vs alpha}.

\begin{table}[ht]
\centering
\caption{Alphanumeric and Oracle OCR Models Comparison}
\small 
\begin{tabular}{|c|c|}
\hline
\textbf{Model Version}                    & \textbf{Accuracy (\%)} \\ \hline
\textbf{Alphanumeric OCR} & 65                  \\ \hline
\textbf{Oracle OCR}                & 71.25                  \\ \hline
\end{tabular}
\label{table: oracle vs alpha}
\end{table}

The Oracle OCR model achieved a higher accuracy of 71.25\%, a modest improvement over the 60\% accuracy obtained with the alphanumeric OCR, which utilizes both the text detection and recognition trained models.

\section{Conclusion and Future Work}

This paper presented the "AI for AR in Satellite AIT" project, demonstrating the integration of AI and AR for enhanced satellite assembly. The system, utilizing Microsoft HoloLens 2, employs computer vision for object detection, 6D pose estimation, and OCR, offering real-time operator guidance.

A key achievement was the successful use of synthetic data for AI model training, addressing the scarcity of real-world datasets in dynamic satellite manufacturing. Experimental results showed high precision in object detection and satisfactory accuracy in 6D pose estimation, despite real-time constraints. The OCR module, particularly for numerical recognition, demonstrated promising performance.  

Moreover, the creation of the SAMAL significantly accelerates the annotation process, enhancing the overall efficacy of the system by up to 20 times.

Future work includes enhancing synthetic data generation, improving 6D pose estimation robustness, developing a complete alphanumeric OCR model, seamless AI module integration, and user evaluation for refined usability. 

\section*{Acknowledgment}

 Thanks to Gonçalo Silva for his contributions in the earlier stages of this research, and to James Pandey and LusoSpace for their support throughout the project.



\bibliographystyle{unsrt}  
\bibliography{biblio}  

\end{document}